# Sentiment Analysis On YouTube Comments Using Machine Learning Techniques Based On Video Games Content

Adi Danish Bin Muhammad Amin
*Faculty of Computing*
*Universiti Malaysia Pahang*
*Al-Sultan Abdullah*
Pahang, Malaysia

Mohaiminul Islam Bhuiyan
*Faculty of Computing*
*Universiti Malaysia Pahang*
*Al-Sultan Abdullah*
Pahang, Malaysia

Nur Shazwani Kamarudin*
*Faculty of Computing*
*Universiti Malaysia Pahang*
*Al-Sultan Abdullah*
Pahang, Malaysia
nshazwani@ump.edu.my

Zulfahmi Toh
*Faculty of Computing*
*Universiti Malaysia Pahang*
*Al-Sultan Abdullah*
Pahang, Malaysia

Nur Syafiqah Nafis
*Faculty of Computing*
*Universiti Malaysia Pahang*
*Al-Sultan Abdullah*
Pahang, Malaysia

*Abstract*—The rapid evolution of the gaming industry, driven by technological advancements and a burgeoning community, necessitates a deeper understanding of user sentiments, especially as expressed on popular social media platforms like YouTube. This study presents a sentiment analysis on video games based on YouTube comments, aiming to understand user sentiments within the gaming community. Utilizing YouTube API, comments related to various video games were collected and analyzed using the TextBlob sentiment analysis tool. The pre-processed data underwent classification using machine learning algorithms, including Naïve Bayes, Logistic Regression, and Support Vector Machine (SVM). Among these, SVM demonstrated superior performance, achieving the highest classification accuracy across different datasets. The analysis spanned multiple popular gaming videos, revealing trends and insights into user preferences and critiques. The findings underscore the importance of advanced sentiment analysis in capturing the nuanced emotions expressed in user comments, providing valuable feedback for game developers to enhance game design and user experience. Future research will focus on integrating more sophisticated natural language processing techniques and exploring additional data sources to further refine sentiment analysis in the gaming domain.

## I. INTRODUCTION

The gaming industry is experiencing unprecedented growth, with projections reaching $300 billion by 2025 [1]. This expansion intensifies competition across gaming sectors, making customer feedback crucial for strategic decisions regarding product development and enhancements [2]. Sentiment analysis (SA) has emerged as a vital tool for real-time monitoring of player feedback across social media platforms, particularly YouTube, Twitter, and Reddit. These platforms provide rich repositories of user opinions that reveal gaming experiences and preferences [3]. YouTube serves as a central hub for gaming content, offering valuable sentiment data that guides industry decision-making and product optimization [1]. While consumers increasingly depend on reviews before purchasing games, traditional critic ratings often differ from public sentiment [3], [4]. Current sentiment analysis approaches face significant challenges in gaming contexts, including comment brevity, subjective language, sarcasm, and spelling errors that hinder accurate interpretation [4], [5]. Additional complications arise from user biases, limited annotated training data, multilingual content, and evolving gaming terminology [5], [6]. These limitations necessitate developing improved sentiment analysis models for video game evaluation based on YouTube comments. Such models enable development teams to understand user acceptance and enhance future design decisions by addressing negative feedback concerns. Effective sentiment analysis provides developers with insights into user preferences, design refinement opportunities, brand reputation monitoring, and competitive positioning. This research ad- dresses these challenges through the following contributions:

- Developing a robust sentiment analysis model using SVM
- Comparative analysis of machine learning models

## II. RELATED WORK

The gaming industry has witnessed exponential growth in recent years, with platforms like YouTube serving as influential spaces for discussions and reviews. Sentiment analysis on video games, particularly based on YouTube comments, offers valuable insights into player experiences, preferences, and trends within the gaming community.

Britto and Pacifico et al. developed a sentiment classification system to determine user approval of video games using Brazilian Portuguese reviews from Steam. Their preprocessing converted text to lowercase and removed special characters, punctuation, and digits. For feature extraction, they employed Bag-of-Words (BoW) to create Document-Term Matrices where columns represent words, rows represent documents, and cells indicate word frequency. Three algorithms were evaluated: Logistic Regression achieved 82.40% accuracy through linear feature combinations, Random Forest attained 79.89% by combining multiple Decision Trees, and Support Vector Machine reached 82.54% using optimal hyperplane separation. Evaluation employed K-Folds Cross Validation with ten-fold partitioning [7]. Zuo et al. conducted sentiment analysis on Steam reviews using Naïve Bayes and Decision Tree classifiers. Data collection utilized Steam Database and web scraping for comprehensive reviews. Preprocessing included special character removal, lowercase conversion, stop word elimination using NLTK, Porter Stemmer application, and frequency threshold filtering. N-gram analysis evaluated unigram, bigram, and trigram models. Classification employed Decision Tree with Information Gain/Gini Index and Gaussian Naïve Bayes with class conditional independence assumptions. Evaluation used 75-25% train test split cross-validation [8]. Tanesab et al. analyzed YouTube comment sentiments using Support Vector Machine on governmental criticism comments. Data collection employed snipping methods from random YouTube videos. Preprocessing removed links, symbols, and stop words, followed by tokenization for word separation. A lexicon-based approach assigned polarity scores (+1 positive, -1 negative, 0 neutral) for sentiment determination. The SVM classifier achieved 84% accuracy on 1000 balanced comments across positive, neutral, and negative classes. Evaluation utilized Confusion Matrix Analysis measuring True Positive, True Negative, False Positive, and False Negative rates [9].

III. METHODOLOGY

This research methodology explores sentiment patterns in YouTube gaming comments using a formal four-phase structure that progresses in serial from data acquisition to performance evaluation. YouTube's Application Programming Interface is utilized in the data collection phase to extract user comments from specific gaming videos and immediately generate preliminary sentiment scores with TextBlob sentiment analysis tools. The data preprocessing phase thereafter implements a variety of text cleaning processes such as lowercase conversion, special character depletion, tokenization, stop word removal, and lemmatization to standardize the data and improve algorithm processing. In the classification phase, preprocessed comments are assigned to positive, negative, and neutral sentiment classes through three separate machine learning algorithms: Naïve Bayes, Logistic Regression, and Support Vector Machine based on the polarity scores obtained via TextBlob sentiment analysis. Finally, in the performance measurement phase, the accuracy scores of the algorithms are calculated and presented visually through sentiment polarity histograms, confusion matrices, and performance comparisons. Figure 1 shows the design of the study. Overall, the presented methodology provides a simple yet effective method of selecting the best performing algorithm with empirical evidence from gaming community comments.

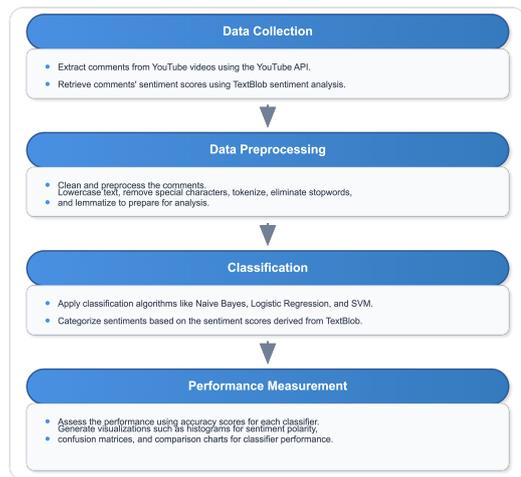

Fig. 1. Study Design

*A. Data Collection*

The sentiment analysis process begins with data extraction using the YouTube API to obtain comments from targeted gaming videos, providing access to user-generated content for analysis. This study collected data from three IGN gaming review videos: Skull Island: Rise of Kong (854 comments), Elden Ring (3,526 comments), and Red Dead Redemption 2 (7,401 comments), totaling 11,781 comments. The research focused exclusively on primary comments while excluding reply threads to maintain consistency in analyzing direct user interactions with gaming content. TextBlob serves as the primary sentiment analysis tool for processing the collected comments [10]. This choice was driven by the study's two-stage methodology where TextBlob provides baseline sentiment scores that are subsequently refined through machine learning classification algorithms. While advanced tools like VADER [11] or BERT-based models might better handle gaming slang and sarcasm, TextBlob's continuous polarity scoring (-1 to +1) provides suitable initial labels for training machine learning classifiers with computational efficiency and reproducibility. The research focuses on comparing traditional ML algorithms (SVM, Naïve Bayes, Logistic Regression) rather than optimizing the initial sentiment assignment tool.

Each comment undergoes assessment to determine its polarity - positive, negative, or neutral-based on linguistic patterns and textual characteristics that indicate emotional significance. Comments are then aggregated into a structured dataset serving as the foundation for subsequent preprocessing and classification steps. This dataset enables identification of gaming community trends, player satisfaction levels, and audience reactions to different game genres and titles, providing valuable insights for game development, design, and marketing strategies.

## B. Data Pre-processing

Before sentiment analysis, data preprocessing is essential to prepare comments for analytical procedures. The preprocessing steps include lowercasing, stop word removal, tokenizaion, lemmatization, special character elimination, and data cleaning. Lowercasing [12] ensures text consistency by treating variations like "Hello" and "hello" as identical terms. This standardization prevents capitalization from distorting sentiment evaluation and enables more reliable algorithm outcomes. Stop words such as "the", "and", "is", and "are" are frequently occurring terms with limited semantic value that can be re- moved without affecting textual meaning [13]. These linguistic function words provide connectivity rather than substantive content, contributing little to sentiment evaluation. Removing these elements reduces dataset noise and enhances algorithm performance by focusing on emotionally significant words. Tokenization involves splitting text into individual words or tokens [14], enabling sentiment analysis algorithms to process words as separate entities. Lemmatization normalizes word variations by converting them to their base or dictionary forms [15]. Special characters including punctuation marks, symbols, and non-alphanumeric characters are eliminated to improve text structure. Data cleaning removes additional whitespace and formatting artifacts from previous processing stages to ensure a clean and standardized dataset.

## C. Data Classification

The dataset was partitioned using an 80-20 split following established machine learning practices [16]. This ratio provides sufficient training data for models to learn underlying patterns while reserving adequate testing data for validation and overfitting detection. The 80-20 split balances training sufficiency with testing adequacy, promoting reproducibility and computational efficiency.

Text classification categorizes textual data into predefined classes based on specific characteristics. For sentiment analysis of YouTube comments, this process includes several phases:

- Feature Extraction: Converting preprocessed text into ma- chine learning-compatible formats using numerical representations like Bag-of-Words or TF-IDF vectors. TF-IDF was selected for its effectiveness in weighting gaming-specific vocabulary while reducing common word impact, providing better sentiment discrimination compared to Bag-of-Words.
- Training and Testing Data Split: Separating the dataset into training sets for model development and testing sets for performance evaluation.
- Model Training: Employing machine learning algorithms including Naïve Bayes, Logistic Regression, and Support Vector Machines to learn text patterns for sentiment classification into positive, negative, and neutral categories.
- Model Evaluation: Assessing trained model performance using testing datasets by comparing predicted sentiment labels with actual labels to measure accuracy and other metrics.
- Sentiment Categorization: Assigning final sentiment labels to YouTube comments based on classification model predictions.

Three machine learning algorithms have used: Naïve Bayes, Logistic Regression and Support Vector Machine.

## D. Naïve bayes

The Naïve Bayes classifier is a probabilistic machine learning model designed for classification tasks [17]. Fundamentally, this classifier operates based on Bayes' theorem, allowing the calculation of the probability of event A occurring given that event B has already taken place. Here, B represents the evidence, while A represents the hypothesis. The crucial assumption in Naïve Bayes is the independence among predictors or features. This assumption suggests that the presence or occurrence of one specific feature doesn't affect another, hence the term "naïve".

$$P\left(\frac{A}{B}\right) = \frac{P\left(\frac{B}{A}\right) \times P(A)}{P(B)} \quad (1)$$

where:

- $P\left(\frac{A}{B}\right)$ represents the probability of event $A$ happening when event $B$ has occurred.
- $P\left(\frac{B}{A}\right)$ stands for the probability of event $B$ occurring given that event $A$ has occurred.
- $P(A)$ and $P(B)$ denote the probabilities of events $A$ and $B$ occurring separately or independently.

## E. Logistic Regression

Logistic regression represents a machine learning approach employed for binary classification problems [18]. This method calculates the probability that a given instance belongs to a specific class based on its input characteristics. Rather than producing continuous outputs as seen in linear regression, logistic regression employs the logistic (sigmoid) function to restrict outputs within the 0 to 1 range, thereby representing probability values.

$$\sigma(z) = \frac{1}{1 + e^{-z}} \quad (2)$$

The logistic function uses a linear combination of input features ($z$) to generate probability outputs between 0 and 1. Model parameters are optimized through gradient descent, which minimizes log-likelihood loss by iteratively adjusting coefficients. This algorithm creates a decision boundary using a threshold (commonly 0.5) to separate classes, where the sigmoid transformation converts continuous inputs into probability scores for binary classification.

## F. Support Vector Machine

Support Vector Machines (SVM) are widely used machine learning algorithms for classification and regression tasks, serving as the primary method in this research [19]. SVM excels in high-dimensional spaces and handles both binary and multiclass problems effectively, though it requires careful parameter tuning and can be computationally intensive for large datasets. SVM operates by finding an optimal hyperplane that separates different classes with maximum margin. The algorithm identifies the decision boundary that maintains the greatest distance from the nearest data points of each class, known as support vectors. This margin maximization strategy ensures clear class separation and improves the algorithm's ability to generalize and make accurate predictions on new data.

## G. Performance Measure

- Accuracy Score: It measures the ratio of correctly classified comments out of the total number of comments in the test dataset. It's a simple and commonly used metric to evaluate classification performance.

$$Accuracy = \frac{TP + TN}{TP + FP + TN + FN} \quad (3)$$

True Positives (TP), True Negatives (TN), False Positives (FP), False Negatives (FN): These components collectively constitute the confusion matrix, offering a more comprehensive perspective on the classification performance. TP and TN represent correct classifications, while FP and FN indicate misclassification.

- Precision: It measures the model's capability to accurately recognize a specific class. Precision can be formulated or defined as follows:

$$Precision = \frac{TP}{TP + FP} \quad (4)$$

- Recall (Sensitivity): It calculates the ratio of actual positives that were accurately predicted by the model. Thus, recall can be formulated as follows:

$$Recall = \frac{TP}{TP + FN} \quad (5)$$

- F1-Score: It is the harmonic mean of precision and recall. It considers both precision and recall, resulting in a lower value compared to accuracy. The formula for calculating the F1-score looks like this:

$$F1 = \frac{2 \times Precission \times Recall}{Precision + Recall} \quad (6)$$

## IV. RESULTS

The sentiment analysis outcomes can demonstrate varying trends, influenced by the size and nature of comments extracted from different videos. This experimentation involved analysing comments from diverse videos with varying comment sizes, aiming to assess the potential impact of comment volume on algorithm performance. The analysis encompassed a selection of videos spanning different genres and engagement levels within the gaming community. This variation in video content contributed to disparate comment sizes, reflecting diverse user interactions and discussions. The sentiment analysis models—utilizing Naïve Bayes, Logistic Regression, and Support Vector Machines (SVM)—were consistently applied across these videos to discern sentiment trends.

### A. Skull Island: Rise of Kong Review

The analysis focused on the IGN video "Skull Island: Rise of Kong Review". The generated data includes the samples of comment extraction paired with their corresponding scoring of polarity and classification of sentiment. Additionally, model evaluation was performed with the support of classification accuracy metrics, and more in-depth model comparison was utilized via detailed metric results of the using algorithms.

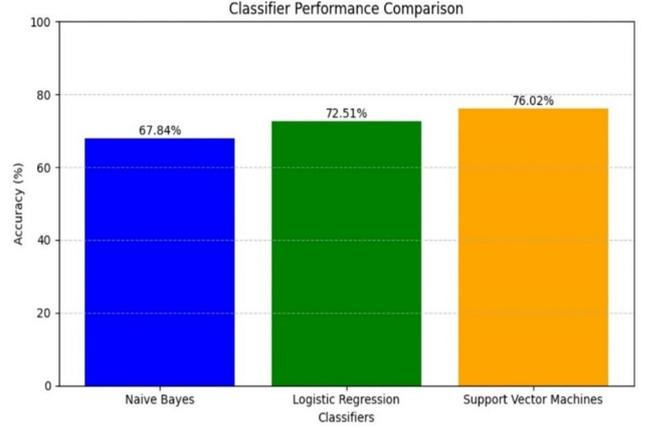
Fig. 2. Accuracy comparison of different models

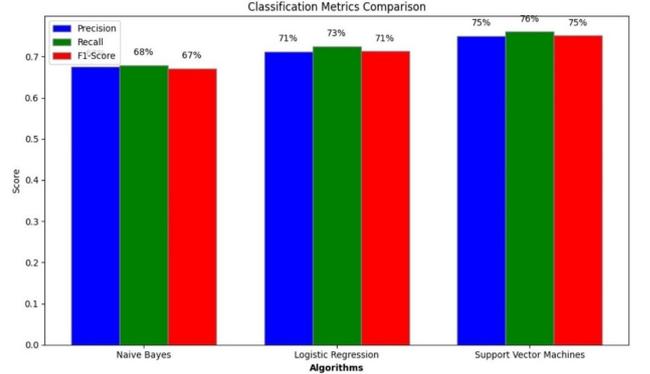
Fig. 3. Comparison of different models on different evaluation matrices

As a result, the analysis showed (figure 2 and 3) that the SVM algorithm demonstrated superior performance for this sentiment analysis task compared to the other evaluated methods. The findings consistently highlighted SVM's effective- ness in handling sentiment classification, yielding the highest classification accuracy across all performance metrics. This superior performance was achieved on the dataset consisting of 854 extracted comments from IGN's "Skull Island: Rise of Kong Review" video, establishing SVM as the most reliable algorithm for this specific gaming content analysis.

### B. Elden ring Review

Furthermore, the sentiment analysis extended to another IGN video, the "Elden Ring Review". The collected data comprised extracted comments paired with their corresponding polarity scores and sentiment classifications. Additionally, classification accuracy evaluations were conducted alongside comprehensive metric comparisons across the employed algorithms.

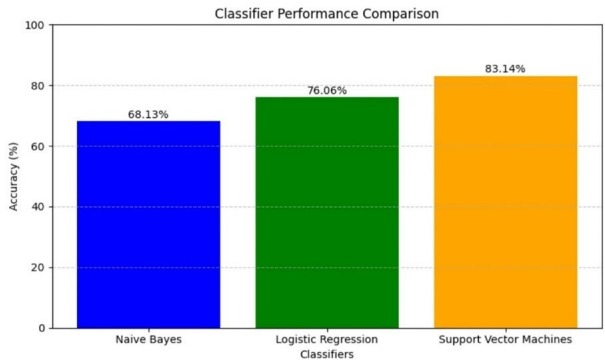

Fig. 4. Accuracy comparison of different models

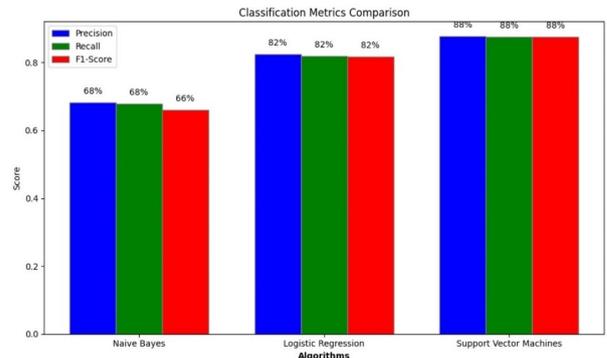

Fig. 7. Comparison of different models on different evaluation matrices

The analysis of 7401 comments from the "Elden Ring Review" video by IGN indicated that (figure 6 and 7) the Support Vector Machine algorithm demonstrated the most superior performance among the employed algorithms in sentiment analysis, achieving superior classification accuracy.

The findings revealed variations in sentiment distribution and classifier performance based on comment volumes. Larger datasets exhibited complex sentiment distributions, while smaller collections showed oversimplified categorization due to limited context. Data extraction focused exclusively on primary comments, omitting replies to ensure concentrated analysis of direct user interactions and avoid diluting sentiment patterns. Performance comparison across datasets showed varying efficiency levels, with Support Vector Machine achieving superior results in all cases. These findings emphasized the importance of considering comment volume in sentiment analysis and confirmed the significance of algorithmic flexibility for accurate sentiment classification in gaming-related YouTube discussions.

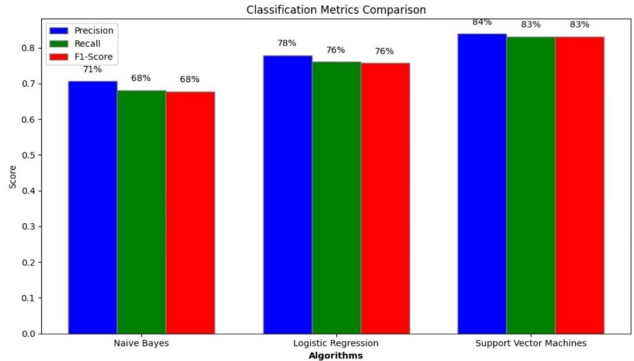

Fig. 5. Comparison of different models on different evaluation matrices

Data set for evaluation contained 3,526 comments retrieved from the video Elden Ring Review of IGN. The results showed (figure 4 and 5) once again the superior performance of the SVM algorithm for sentiment analysis, as it demonstrated the highest classification accuracy rating. This mirrors the results retrieved for the 854-comment dataset from the Skull Island: Rise of Kong Review, which supports the effectiveness of SMV in conducting sentiment classification.

*C. Red Dead Redemption 2 Review*

Extending the sentiment analysis to incorporate the "Red Dead Redemption 2 Review" video from IGN, an extensive dataset was assembled. The dataset comprised extracted comments, their associated polarity values, and classified sentiments. Furthermore, a thorough comparison of metrics was performed among the range of algorithms utilized in this investigation.

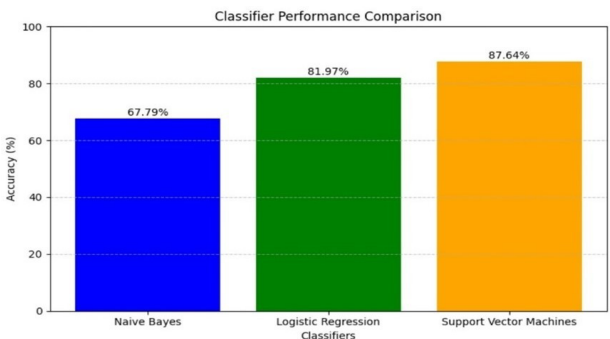

Fig. 6. Accuracy comparison of different models

## V. CONCLUSION AND FUTURE WORK

The sentiment analysis of YouTube comments on gaming videos produced an insightful body of user sentiments existing within the gaming community. Evaluating different videos and industry sources, like IGN's channel, ultimately proved the varying tones and attitudes in viewers' reactions. The re- search used diverse machine learning algorithms, proving their functionality in capturing sentiment patterns across distinct volumes of comments and video types. At the same time, there were several limitations and other constraints that can be viewed in the body of linear.

The most significant limitation was based on the disparity in the ability to access comments. In some videos, comments were well below the 500 marks, while the volume in others exceeded several hundreds of thousands, which somewhat impacted algorithm performance. Furthermore, YouTube comments specifics, combined with the frequent use of informal or ironic language and unclear formulations, added complexity to categorizing sentiments precisely. Similarly, despite the effectiveness of algorithms, they seem to lack the potential to capture all the nuances of user sentiments based on a comment, which remains short for sentiment context. The study's reliance on TextBlob for sentiment analysis limits its ability to capture sarcasm and gaming-specific language common in YouTube comments. The research used only traditional machine learning algorithms (Naïve Bayes, Logistic Regression, SVM) without exploring deep learning approaches like LSTM or BERT that could better understand contextual relationships. Feature extraction was restricted to TF-IDF without considering contextual embeddings such as Word2Vec or GloVe that might provide richer semantic understanding. The research opens several opportunities for further and more advanced sentiment analysis of YouTube gaming comments. The use of more sophisticated natural language processing could aid in a more nuanced recognition of algorithms' emotional tone. Besides, creating models sensitive to the context of sentiment could facilitate the determination of the interpretation of nuanced sentiments typical for gaming comments, gaming slang, emojis, sarcasm. Temporal analysis, particularly assessing patterns of commentary over time, including game releases or major changes and developments across the industry, can provide a new perspective on sentiment patterns. Other data sources, such as gaming platforms and social media, could be included in the analysis to obtain a more desired sentiment picture transcending video comment. Therefore, the research outlines patterns in sentiment of gaming communities on YouTube, providing the groundwork for further analysis and more general research in the future on sentiment methodologies and points of interests of sentiment in YouTube gaming comments.

## ACKNOWLEDGMENT

This work is supported by Fundamental Research Grant Scheme 2024 by Ministry of Higher Education Malaysia FRGS/1/2024/ICT02/UMP/02/3 (RDU240125) and partially supported by UMPSA Research Grant (Grant No. RDU230353).